\documentclass[journal]{IEEEtran}

\usepackage[T1]{fontenc} 
\usepackage{amsmath,amssymb,nicefrac}
\usepackage{txfonts}
\usepackage{bm} 
\usepackage[compress,noadjust]{cite}
\usepackage{hyperref}
\usepackage{graphicx}
\usepackage{tikz}
\usetikzlibrary{shapes.geometric, arrows,fit}
\usepackage{placeins}
\usepackage{float}
\usepackage{booktabs}
\usepackage{multirow}

\usepackage{calc}

\usepackage{chngcntr}
\counterwithin{paragraph}{section}

\def\vector#1{{\boldsymbol{#1}}}
\def\tablescale{.98}	
	
\begin{document}
\bstctlcite{IEEEexample:BSTcontrol}

\author{J.~Hannink, T.~Kautz, C.~F.~Pasluosta, K.-G. Ga\ss mann, J.~Klucken,  B.~M.~Eskofier,~\IEEEmembership{Member,~IEEE,~EMBS}
	
	\thanks{J. Hannink, T. Kautz, C. F. Pasluosta and B. M. Eskofier: Digital Sports Group, Pattern Recognition Lab, Department of Computer Science, University of Erlangen-N\"urnberg (FAU), Germany}
	\thanks{K.-G. Ga\ss mann: Geriatrics Centre Erlangen, Waldkrankenhaus St. Marien, Erlangen, Germany}
	\thanks{J. Klucken: Department of Molecular Neurology, University Hospital Erlangen, University of Erlangen-N\"urnberg (FAU), Germany}
	\thanks{Corresponding author: J. Hannink, \href{mailto:julius.hannink@fau.de}{julius.hannink@fau.de}}

}

\title{Sensor-based Gait Parameter Extraction with Deep Convolutional Neural Networks}

\maketitle

\begin{abstract}
Measurement of stride-related, biomechanical parameters is the common rationale for objective gait impairment scoring. State-of-the-art double integration approaches to extract these parameters from inertial sensor data are, however, limited in their clinical applicability due to the underlying assumptions.

To overcome this, we present a method to translate the abstract information provided by wearable sensors to context-related expert features based on deep convolutional neural networks. Regarding mobile gait analysis, this enables integration-free and data-driven extraction of a set of eight spatio-temporal stride parameters. To this end, two modelling approaches are compared: A combined network estimating all parameters of interest and an ensemble approach that spawns less complex networks for each parameter individually.

The ensemble 
approach is outperforming the combined modelling in the current application. On a clinically relevant and publicly available benchmark dataset, we estimate stride length, width and medio-lateral change in foot angle up to $\boldsymbol{-0.15\pm6.09}$ cm, $\boldsymbol{-0.09\pm4.22}$ cm and $\boldsymbol{0.13 \pm 3.78^\circ}$ respectively. Stride, swing and stance time as well as heel and toe contact times are estimated up to $\boldsymbol{\pm 0.07}$, $\boldsymbol{\pm0.05}$, $\boldsymbol{\pm 0.07}$, $\boldsymbol{\pm0.07}$ and $\boldsymbol{\pm0.12}$ s respectively. This is comparable to and in parts outperforming or defining state-of-the-art.

Our results further indicate that the proposed change in methodology could substitute assumption-driven double-integration methods and enable mobile assessment of spatio-temporal stride parameters in clinically critical situations as e.g. in the case of spastic gait impairments.
\end{abstract}

\begin{IEEEkeywords}
deep learning, convolutional neural networks, regression, mobile gait analysis, spatio-temporal gait parameters
\end{IEEEkeywords}

\section{Introduction}
A variety of neurological and musculoskeletal diseases affect human gait quality and manifest in specific stride characteristics. Parkinson's disease (PD), for example, is associated with a reduced stride length, shuffling steps or impaired gait initiation. As reduced gait quality can lead to severe reductions in patient mobility and quality of life \cite{Ellis2011}, it is important to quantify, detect and treat gait impairments as early as possible.

Objective quantification of gait impairment is based on stride-specific characteristics such as stride length or stride time. These parameters are commonly extracted with the help of several electronic measurement systems including computerised pressure mats (\cite{Givon2009,Webster2005}), optical motion-capture systems \cite{Kressig2004} or mobile, sensor-based solutions (\cite{Klucken2013,Mariani2013,Rampp2015,Ferrari2015,Trojaniello2014,Aminian2002,Salarian2013}). While the first two require a laboratory environment and are limited in availability, the latter is mobile and inexpensive. This renders mobile, sensor-based solutions the primary choice for unobtrusive gait analysis systems.
 
Choosing this modality though introduces a conflict between the abstract variables of measurement and the readout parameters requested by the users and is as such entangled with the physical constraints in wearable sensing: For instrumented medical healthcare applications, one might be able to measure accelerations and angular rates at a patient's foot using state-of-the-art inertial sensors. However, the treating physician is not interested in interpreting acceleration signatures for a given stride but rather wants to monitor variables directly related to the situation as for example stride length or heel-strike angle. The efficient translation of abstract data to context-related knowledge thus is the underlying challenge in all applications of wearable sensors and mobile healthcare technologies.

In the extraction of stride parameters, this challenge is addressed from several perspectives. The majority of methods are based on physical and geometric reasoning to extract spatial gait parameters using double-integration of inertial sensor signals (\cite{Mariani2013,Rampp2015,Rebula2013,Ferrari2015,Trojaniello2014}). The main limitation regarding this type of approach is the dependency on a zero-velocity phase within each stride that is needed to re-initialize the integration process. In clinical practice, however, this assumption is easily violated \cite{Hannink2016}. Other approaches aim at driving bio-mechanical models of the lower extremity with sensor data (\cite{Aminian2002,Salarian2013}) or apply machine learning approaches in order to extract the parameters of interest \cite{Hannink2016}. 

The underlying problem of efficient data to knowledge translation is recently being addressed very successfully in the field of image understanding. Here, data from images is identified to belong to a certain object class \cite{He2015}, translated to captions that describe the image content \cite{Vinyals2015} or used to identify persons based on face recognition \cite{Taigman2014}. All these applications represent ground-breaking advances in their respective field in terms of recognition rates. The common underlying methodology that allows these achievements is a branch of machine learning called deep learning.

Due to its success in other domains, deep learning is starting to appear in the context of wearable sensing and computing to extract meaningful information from sensor data (\cite{Hammerla2015,Hammerla2016,Ordonez2016,Hannink2016,Zheng2014}). This particular branch of machine learning is said to have large potential in mobile sensing and computing regarding inference accuracy, robustness or class-scaling which are partly missing from state-of-the-art \cite{Lane2015}. {Applications of deep learning in wearable sensing and computing are, however, largely focussing on activity recognition }(\cite{Hammerla2015,Hammerla2016,Ordonez2016,Zheng2014}). {To the} {authors' best knowledge, this and their prior work }\cite{Hannink2016} {are the first applications regarding other topics in the field}.

In this work, we present a framework  based on deep convolutional neural networks and aim at translating the abstract information provided by wearable sensors to context-related expert features requested by the users. The system is trained on a regression task between sensor data and a set of reference output parameters. {Thereby, it extends the authors' prior work} \cite{Hannink2016} {that only addresses a single output parameter.} A pre-requisite for this is a knowledge base, i.e. a collection of wearable sensor data captured in a controlled environment and annotated with the help of a reference system that can directly measure the expert features of interest.

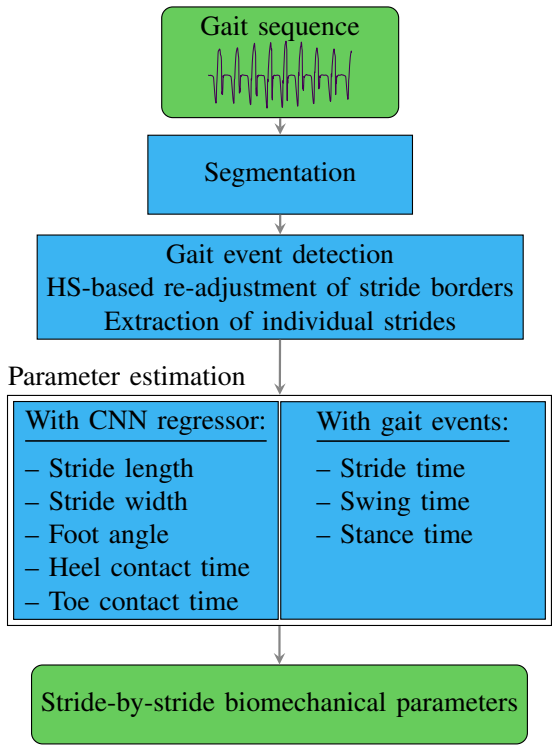
\begin{figure}[!h]
\definecolor{mygreen}{RGB}{103,204,92}
\definecolor{mygray}{RGB}{133,133,133}
\definecolor{myblue}{RGB}{59,180,242}
\centering
\scalebox{1.05}{
\begin{tikzpicture}[node distance=1.6cm]
\tikzstyle{box} = [rectangle, rounded corners, minimum width=3cm, minimum height=1cm,align=center, draw=black, fill=mygreen,anchor=north]
\tikzstyle{process} = [rectangle, minimum width=3.35cm, minimum height=1cm, draw=black, fill=myblue, align=center,anchor=north]
\tikzstyle{arrow} = [thick,->,>=stealth,color=mygray]

\node (gaitsequence) [box] {Gait sequence\\\includegraphics[width=.1\textwidth]{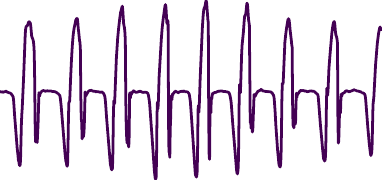}};
\node (segmentation) [process,below of = gaitsequence,yshift=.5em] {Segmentation};
\node (HSHS) [process,below of=segmentation,yshift=.5em]{Gait event detection\\ HS-based re-adjustment of stride borders\\Extraction of individual strides};

\coordinate[below of=HSHS] (c);

\node (cnn) [process,left,align=left,left of =c,yshift=-3.5em,xshift=-.1cm] {\underline{With CNN regressor:}\\[.5em]
												{\textendash} Stride length\\ 
												{\textendash} Stride width\\
												{\textendash} Foot angle\\
												{\textendash} Heel contact time\\
												{\textendash} Toe contact time};
\node (timings) [process,align=left,right of = cnn, right,xshift=.1cm] {\underline{With gait events:}\\[.5em] 
												 {\textendash} Stride time\\
												 {\textendash} Swing time\\ 
												 {\textendash} Stance time\\
												 \null\\
												 \null};

\node (params) [process, fit = (cnn)(timings),inner sep=.2em,fill=none,
				label={[xshift=-5.5em]Parameter estimation}] {};
\node (output) [box,below of = params,anchor=north,yshift=-1em] {Stride-by-stride biomechanical parameters};

\draw [arrow] (gaitsequence) -- (segmentation);
\draw [arrow] (segmentation) -- (HSHS);
\draw [arrow] (HSHS) -- (params);
\draw [arrow] (params) -- (output);
\end{tikzpicture}
}
		\caption{Conceptual flowchart for applying the proposed system to mobile gait analysis: An in-comming gait sequence is segmented and gait events are identified within each segment. In a second step, individual strides are defined from heel-strike (HS) to heel-strike and fed to the biomechanical feature extraction routines. These either compute the timings directly related to the identified gait events or estimate spatio-temporal parameters with a convolutional neural network (CNN).}
		\label{fig:flowchart}
\end{figure}

We apply the proposed framework in the context of mobile gait analysis as illustrated in Fig. \ref{fig:flowchart}. In doing so, we focus specifically on the extraction of biomechanical stride parameters with convolutional neural network regression while potential benefits from deep learning approaches to other parts of the pipeline (e.g. segmentation) might be addressed in future work. A total of eight exemplary and stride-specific characteristics are extracted that are clinically relevant as they define gait quality. To this end, two different modelling approaches are compared: A combined model that uses one network architecture to estimate all expert features of interest and an ensemble approach where one neural network is spawned for each output parameter individually. Both models are trained and evaluated on a publicly available and clinically relevant benchmark dataset. 

In summary, our main contributions are: (1) A generalisable method for data-driven and integration-free extraction of spatio-temporal gait characteristics, and (2) Technical validation of the proposed method on a clinically relevant and publicly available dataset.

\section{Methods}
\subsection{Data Collection and Setup}

We use a benchmark dataset collected by Rampp et al. \cite{Rampp2015} that is publicly available at \url{https://www5.cs.fau.de/activitynet/benchmark-datasets/digital-biobank/} and briefly described here.

\begin{figure}[!h]
		\centering
		\includegraphics[width=.6\columnwidth]{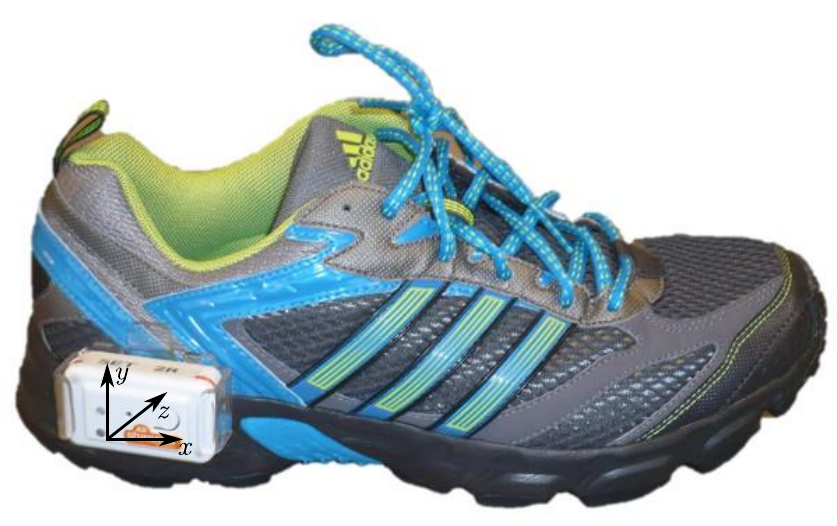}
		\caption{Placement of the inertial sensor and axes definition.}
		\label{fig:shoes}
\end{figure}

The inertial sensor platform Shimmer2R \cite{Burns2010} consisting of a 3d-accelerometer (range $\pm 6\,\text{g}$) and a 3d-gyroscope (range $\pm 500\, ^\circ$/s) was used for data collection. It was attached laterally below each ankle joint (Fig. \ref{fig:shoes}). In order to avoid gait changes due to different shoe characteristics \cite{Menant2009}, the same shoe model (adidas Duramo 3) was used by all subjects. Data was captured at $102.4$ Hz at a resolution of $12$ bit. Simultaneously, validation data was acquired with the well established pressure mat GAITRite with a spatial resolution of $\pm 1.27$ cm \cite{Webster2005}.
\FloatBarrier

In total, 116 geriatric inpatients were assessed at the Geriatrics Centre Erlangen (Waldkrankenhaus St. Marien, Erlangen, Germany). Written informed consent was obtained prior to the gait assessment in accordance with the ethical committee of the medical faculty at Friedrich-Alexander University Erlangen-N\"urnberg (Re.-No. 4208).

\begin{figure*}[!t]
\centering
\includegraphics[width=0.98\textwidth]{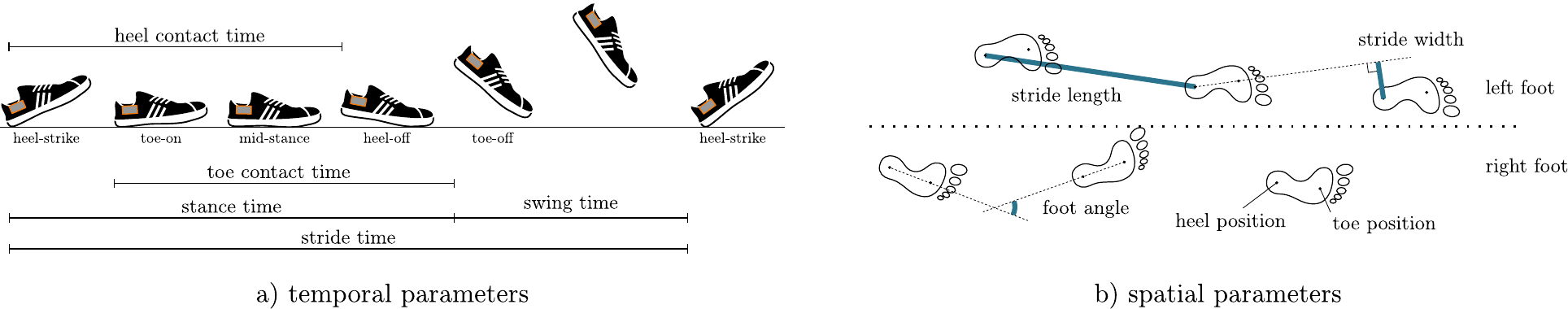}
\caption{Definition of temporal and spatial gait parameters based on heel and toe positions and timings measured with the GAITRite reference system.}
\label{fig:gaitrite_parameters}
\end{figure*}

For our study, the annotation on the dataset was extended with additional parameters compared to the annotation reported on by Rampp et al. \cite{Rampp2015}. This was based on positions and timings of the patients' heel and toe as measured by the GAITRite reference system. The spatial parameter set was enlarged to cover not only stride length, but also stride width and change in medio-lateral foot angle. Additionally, heel and toe contact times were added to the list of temporal parameters stride, stance and swing time.  Fig. \ref{fig:gaitrite_parameters} gives an overview on the definitions of temporal and spatial parameters. Stride width was defined as shown in Fig. \ref{fig:gaitrite_parameters} b) and positive values were measured towards the lateral side of the shoe.

Patients performed an extensive geriatric assessment, described by Rampp et al. \cite{Rampp2015} in detail. For the scope of this paper, we focused on the free walking test over the GAITRite mat at comfortable walking speed instrumented with the inertial sensors. After excluding datasets from eight patients due to medical reasons (i.e. patients could not complete the measurement protocol), two due to inertial sensor malfunction and additional seven due to measurement errors with the GAITRite system, 99 patients were left for training and evaluation of the proposed method. Compared to Rampp et al. \cite{Rampp2015}, reference values for heel and toe contact times could not be computed for two patients.

Gait disorders or fall proneness were diagnosed in 54\% of the study population. The other top three diagnoses were heart rhythm disorder (70\%), arterial hypertension (69\%) and coronary artery disease (41\%), which are also associated with gait and balance disorders \cite{Salzman2010}. In summary, this dataset constitutes a clinically relevant study population both in terms of the number of patients and the presence of unpredictable gait alterations.

{TABLE} \ref{tab:datasetStats} gives an overview on the extended set of reference parameters on the dataset and their mean value, standard deviation, as well as their minimal/maximal values.
\begin{table}[!h]

	\caption{Extended set of annotation parameters on the dataset and their mean values, standard deviations, as well as minimal / maximal values.}
	\centering
	\scalebox{\tablescale}{
	\begin{tabular}{llrr}
		\toprule
		Output Parameter & Unit & Mean $\pm$ Std.\hspace{1em} & [\hfill Min,\hfill Max\hfill] \\\toprule
Stride length & cm & 80.63 $\pm$ \hspace{0em} 23.23 & [\hspace{.3em}20.01,\hspace{0em} 129.81] \\ 
Stride width & cm & -1.44 $\pm$ \hspace{0em} 13.29 & [\hspace{0em}-37.52,\hspace{.5em}  33.03] \\ 
Foot angle & $^\circ$ &  0.07 $\pm$ \hspace{.5em}  3.49 & [\hspace{0em}-11.93,\hspace{.5em}  15.86] \\ \midrule
Stride time & s &  1.23 $\pm$ \hspace{.5em}  0.19 & [\hspace{.58em} 0.74,\hspace{1em}   2.06] \\ 
Swing time & s &  0.37 $\pm$ \hspace{.5em}  0.08 & [\hspace{.58em} 0.01,\hspace{1em}   1.05] \\ 
Stance time & s &  0.85 $\pm$ \hspace{.5em}  0.16 & [\hspace{.58em} 0.48,\hspace{1em}   1.65] \\ 
Heel contact time & s &  0.64 $\pm$ \hspace{.5em}  0.14 & [\hspace{.58em} 0.16,\hspace{1em}   1.52] \\ 
Toe contact time & s &  0.69 $\pm$ \hspace{.5em}  0.17 & [\hspace{.58em} 0.25,\hspace{1em}   1.57] \\ 
		\bottomrule
	\end{tabular}
	}
	\label{tab:datasetStats}

\end{table}

\FloatBarrier
\subsection{Preprocessing}
Before the inertial sensor data is fed to the convolutional neural network, we perform a series of preprocessing steps. {The segmentation step mentioned in the overview } Fig. \ref{fig:flowchart} {is already provided by the dataset in our case. Preprocessing therefore includes} extraction of annotated strides from the continuous recordings, calibration from raw sensor readings to physical units, coordinate system transformations to align sensor axes on left and right feet, normalisation w.r.t. sensor ranges and padding to fixed length of 256 samples per stride to ensure fixed size input to the network. 
\begin{figure}[h]
		\centering
		\includegraphics[width=.98\columnwidth]{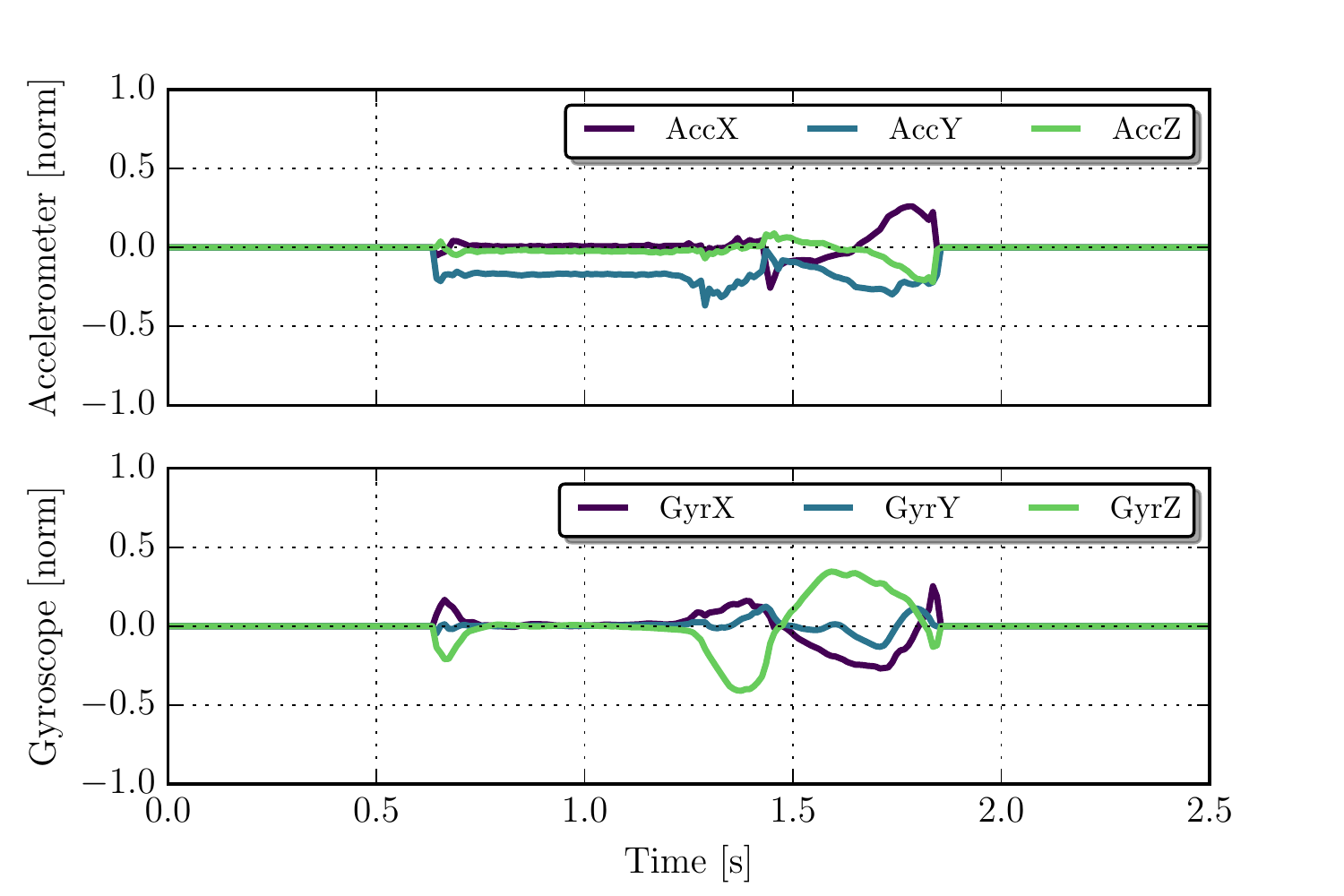}
		\caption{Exemplary input signal for a stride defined from HS$\to$HS after preprocessing.}
		\label{fig:inputSignal}
\end{figure}

The system is trained on data segments from heel-strike to heel-strike (HS). This choice of stride definition is beneficial since it does not assume a zero-velocity phase that state-of-the-art double-integration approaches need to re-initialize the integration process. In clinical practice, {this assumption is easily violated as e.g. in the case of spastic gait impairments.} {However, initial ground contact can still be detected for e.g. spastic gait patterns and provides a valid segmentation of the signal into strides. This is the type of scenario that we intend to address with stride segments defined from HS$\to$HS.}

In order to provide this kind of input data, we need to detect HS and toe-off (TO) events in the stride segmentation provided by the dataset and adjust the stride borders accordingly. Detection of HS and TO events within the sensor data is done according to Rampp et al. \cite{Rampp2015}. An exemplary input signal for one stride defined from HS$\to$HS is shown in Fig. \ref{fig:inputSignal}.

Furthermore, we can directly compute three expert-features from the list in {TABLE} \ref{tab:datasetStats} based on the HS and TO events in the datastream. Stride time is defined as the HS$\to$HS distance and given a TO event, we can subdivide each stride into its two phases and compute stance and swing time {(see Fig. }\ref{fig:gaitrite_parameters}{ a) or Rampp et al. }\cite{Rampp2015}). This leaves a set of five output parameters to be estimated by the deep convolutional neural networks.

\FloatBarrier
\subsection{Network Architectures}
\begin{figure*}[!t]
		\centering
		\includegraphics[width=.98\textwidth]{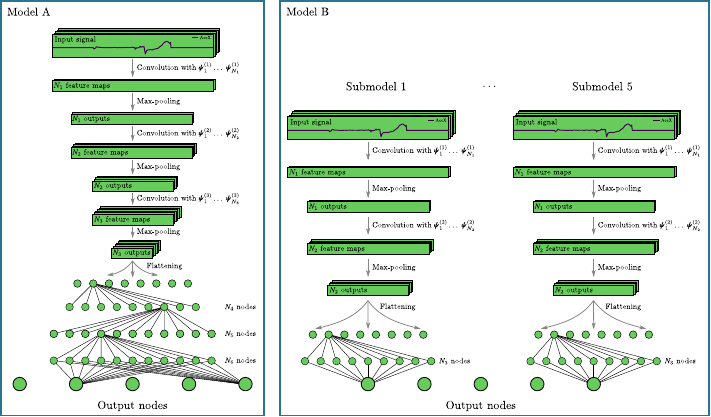}
		\caption{The two network architectures for data to knowledge translation in the context of mobile gait analysis: Model A consists of three convolutional layers with max-pooling followed by three densely connected layers. Model B, however, spawns smaller networks consisting of two convolutional layers with max-pooling and one densely connected layer for each of the output variables. {Additionally, dimensionalities of important layers are indicated.}}
		\label{fig:modelAB}
\end{figure*}

The network architectures used here are based on three elementary building blocks: Convolutional, max-pooling and densely connected layers.

A convolutional connection between layers is defined by a set of $N_i$ kernels $\vector\psi_1 \dots\, \vector\psi_{N_i}$ of length $L_i$ and biases $b_1 \dots \, b_{N_i}$. The index $i$ thereby represents a label for the layer at hand. Given a multi-channeled input vector $\vector{x}_j$ with $j=1...N_{i-1}$, the activation or output of the convolutional connection is computed as
\begin{align}
	\vector{a}_k = ReLU\left(\sum_j \vector\psi_{k,j} * \vector{x}_j + b_k\right)
	\label{eq:convLayer}
\end{align}
with $k=1\dots N_i$. We used a rectifying linear unit ($ReLU$) as the activation function for this type of connection {as this non-linearity is said to model biological neurons better compared to tangent or sigmoid functions} \cite{Glorot2011}.

Convolutional layers are often followed by max-pooling layers to increase robustness of the extracted features \cite{Boureau2010}. This type of connection downsamples the feature maps obtained by the convolutional connection by taking the maximum in temporal windows of length $r$. The downsampling factor is thus $1/r$.

The third type of connection used here is the densely connected layer. This type of connection is defined by a set of weight vectors $\vector{W}_1 \dots \vector W_{N_i}$ and biases $b_1 \dots b_{N_i}$. Given a single-channel input vector $\vector{x}$, the activation of the densely connected layer is computed by matrix-multiplication as
\begin{align}
	a_j = ReLU\left(\sum_i W_{i,j} \, x_i + b_j\right)
	\label{eq:FullLayer}
\end{align}
with $j = 1\dots N_i$. In case the output of the previous layer is a multi-channelled vector, the single-channel input vector $\vector x$ is constructed by concatenation of the individual channels i.e. flattening. Again, we use a rectifying linear unit for activation.

Finally, a readout layer compresses the last dense layer to the number of output variables for the task at hand. The readout layer is identical to a densely connected layer with the identity instead of the $ReLU$ as an activation function. The number of output variables is then encoded in the number of weight vectors for this layer.

Based on these elementary blocks, two models are built:
\begin{itemize}
	\item Model A: Estimating the complete set of output variables with a combined model (Fig. \ref{fig:modelAB}, left).
	\item Model B: Estimating each output variable individually with an ensemble of networks (Fig. \ref{fig:modelAB}, right).
\end{itemize}
Consequently, the individual network architectures in model B can be less complex compared to model A in order to achieve comparable model complexities.

Regarding the application of data to knowledge translation in the context of mobile gait analysis, we decide for a network architecture built from three convolutional layers with max-pooling followed by three densely connected layers and a readout layer for model A. In the convolutional layers we train $N_1 = 32,\, N_2 = 64 $ and $ N_3 = 128 $ kernels of size $30,15$ and $7$ samples respectively as well as the corresponding number of bias terms. Max-pooling is done in non-overlapping, temporal windows of size $r=2$ samples. Given the sampling frequency, the kernel size corresponds to approximately $0.29$ s on all three layers. The three densely connected layers are trained with $N_4=2048,\, N_5=1024$ and $N_6=512$ weight vectors and bias terms respectively. The readout layer has $N_\text{output} = 5$ nodes for model A. 

For model B, the individual network architectures are built from two convolutional layers with $N_1 = 16$ and $N_2 = 32$ filters of size $30$ and $15$ samples respectively and one densely connected layer with $N_3 = 1024$ nodes. The max-pooling layers are identical to model A with a downsampling factor of $\nicefrac{1}{2}$. The readout layer, however, has only $N_\text{output} = 1$ node as each individual architecture is responsible for one of the output parameters. {Fig.} \ref{fig:modelAB} {gives an overview on the network architectures used in both models.}

{The theoretical motivation for this choice is to address the most crucial question in network design of global vs. individual modelling with two representative cases. In model A, we only distinguish between different kinds of output parameters at the last level of the network. The features extracted in this architecture therefore have to be general enough to capture information about all of the output parameters. In model B, however, each output parameter has its own feature extraction path that can be optimized to the parameter at hand.}

\subsection{Training}

Training of neural networks is viewed as an optimization problem regarding an error function (implicitly) depending on the network parameters. This error defines a discrepancy measure between predicted output and a ground truth reference on the training dataset or subsets thereof. Weights and biases on all layers are then changed using back-propagation and with the aim to minimize the error. In practice, however, only random subsets of the training dataset (mini-batches), are shown to the optimizer in one iteration of the training loop to speed up the learning phase (stochastic learning) \cite{Goodfellow2016}.

Because of the different numeric ranges and physical units in the output parameters (see {TABLE} \ref{tab:datasetStats}), the network is trained to estimate normalized and dimensionless output variables $\hat y_i$. Therefore, each reference $y_{i,\text{ref}}$ is scaled to the range $[0,1]$ using the minimum/maximum value attained on the entire training set $\mathcal{S}_\text{train}$:

\begin{align}
	\hat{y}_{i,\text{ref}} = \frac{y_{i,\text{ref}} - \min_{\mathcal{S}_\text{train}} y_{i,\text{ref}}}{\max_{\mathcal{S}_\text{train}} y_{i,\text{ref}} - \min_{\mathcal{S}_\text{train}} y_{i,\text{ref}}} 
	\label{eq:scaling}
\end{align}
Predictions on the test set are later rescaled to their physical dimensions using the scaling parameters from the training set.

Given predictions $\hat y_i$ on a mini-batch of size $N_\text{batch}$ for each output variable $i=1\dots N_\text{output}$, we define the error as the sum of the individual root-mean-square errors on the mini-batch:

\begin{align}
	E = \sum_i \text{rmsq}\left(\hat y_i - \hat y_{i,\text{ref}}\right)
	\label{eq:error}
\end{align}

For optimization, we use Adam \cite{Kingma2015}, a state-of-the-art optimization method for stochastic learning. On benchmark datasets, it shows faster convergence than other stochastic optimization routines and we use default settings of $\alpha = 1\text{e}^{-3},\, \beta_1=0.9,\,\beta_2 = 0.999$ and $\epsilon=1\text{e}^{-8}$ (for details see \cite{Kingma2015}).
All weights are initialized by sampling a truncated normal distribution with standard deviation 0.01 and biases are initially set to 0.01. We train for a fixed number of 4000 iterations with a mini-batch size of $N_\text{batch}=100$ strides.


To prevent over-fitting, dropout is used on the densely connected layers. This technique effectively samples a large number of thinned architectures on the hidden layers by randomly dropping nodes during training. With this, over-fitting could be significantly reduced in many use-cases and was superior to weight-regularisation methods \cite{Srivastava2014}. We use fixed dropout probabilities of $p^{(4)} = 0.75$, $p^{(5)} = 0.5$ and $p^{(6)} = 0.0$ for model A. For the individual architectures in model B, a dropout probability of $p^{(3)} = 0.5$ is used. Every connection thus has a 50\% chance of being inactive. During testing, however, the full architectures are used and no connections are dropped.

The networks are implemented and trained using google's TensorFlow library \cite{tensorflow2015-whitepaper}.
\FloatBarrier
\subsection{Evaluation Scheme}

Evaluation of the two modelling approaches is based on a 10-fold cross validation scheme. The stride-specific sensor data from 99 patients on the dataset are sorted into training and test partitions depending on the patient identifier to ensure distinct splits of the dataset. 
For each of the two models, we iterate over the complete dataset in this fashion and estimate the output variables on the test set in each fold. The estimates from individual folds are then pooled to arrive at average statistics for each output variable and model. As an evaluation statistic, we use average accuracy $\pm$ precision which correspond to the mean and standard deviation of the signed error distribution. The two models are compared based on this statistic and a Levene test of equal variances between the respective distributions to check whether precisions differ significantly. Because the error distributions for the parameter heel contact time are slightly non-gaussian (checked by visual inspection of q-q plots), a Levene test is preferred over e.g. a Bartlett test that is less robust against non-normality. 

In order to assess the learning speed and performance of model A and B, we compute the training error for each of the models over the training iterations for an exemplary and patient-wise 90/10\% train/test split of the dataset.

\FloatBarrier
\section{Results}
\subsection{Training}
\begin{figure}[!h]
		\centering
		\includegraphics[width=.98\columnwidth]{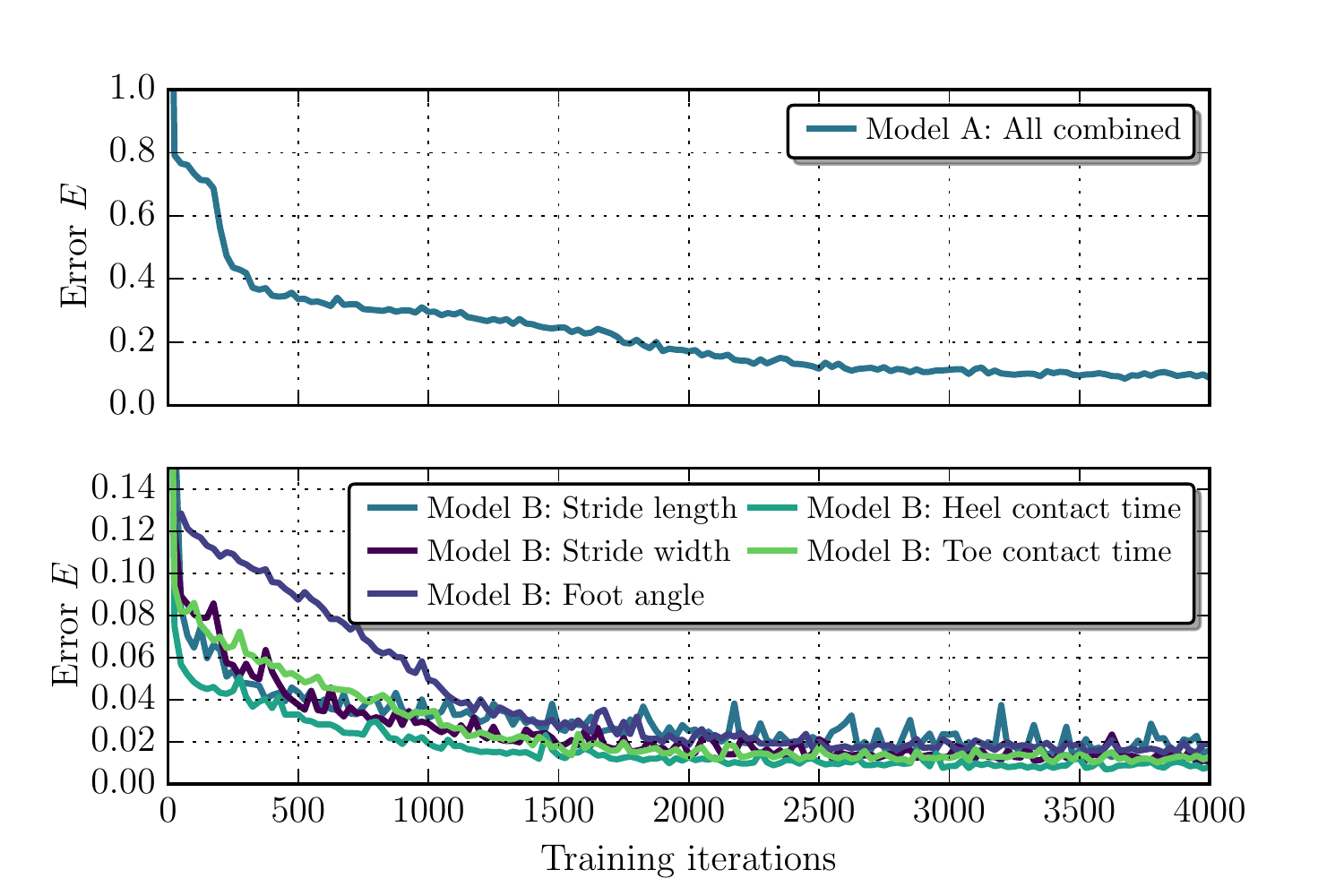}
		\caption{Error $E$ evaluated on the entire training set as a function of the training iterations for model A, all submodels that constitute model B and
		 an exemplary 90/10\% split of the dataset.} 
		\label{fig:trainingError}
\end{figure}

Fig. \ref{fig:trainingError} shows the error evaluated on the entire training set over the iterations for an exemplary 90/10\% train/test split of the dataset. The error is evaluated for model A and for each of the submodels that constitute model B. In all cases, the fixed number of 4000 iterations is sufficient to reach a stable regime of the error on the entire training dataset and hence we stop the training. {Furthermore, the adaptation of the two models to the training data is comparable w.r.t. the selected error function as $E_{1\dots5} \approx 0.02$ in model B corresponds to a total/summed error of $E \approx 0.1$ for model A} (Fig \ref{fig:trainingError}).

\begin{table*}[!t]

\caption{{Comparison of model A and B regarding average accuracy and precision reached on unseen test data. To compare average precisions, a Levene-test was performed on the respective error distributions at the 0.01 significance level.}}
	\centering
	\scalebox{\tablescale}{
	\begin{tabular}{lrrrrr}
		\toprule
		& \multicolumn{1}{c}{Stride length}& \multicolumn{1}{c}{Stride width} & \multicolumn{1}{c}{Foot angle} & \multicolumn{1}{c}{Heel contact time} & \multicolumn{1}{c}{Toe contact time} \\\toprule
		Model A& $-0.34 \pm 8.10 $ cm \
          & $0.41 \pm 7.79 $ cm\
          & $-0.05 \pm 3.59 $ $^\circ$\
          & $-0.00 \pm 0.08 $ s\
          & $-0.00 \pm 0.12 $ s \\ 
        Model B  & $\boldsymbol{-0.15 \pm 6.09} $ \textbf{cm} \
          & $\boldsymbol{-0.09 \pm 4.22 }$ \textbf{cm}\
          & $0.13 \pm 3.78 $ $^\circ$\
          & {$0.00 \pm 0.07$ s}\
          & $0.00 \pm 0.12 $ s \\\midrule 
       {Levene-test}  & sign. & sign.& n.s. & {n.s.} & n.s.\\
		\bottomrule
	\end{tabular}
	}
	\label{tab:results-modelcomparison}
\end{table*}

\subsection{Stride Parameter Estimation on Unseen Data}
{TABLE} \ref{tab:results-modelcomparison} lists average accuracy and precision {on the unseen test data} achieved by the two models w.r.t. the pooled estimates from each cross-validation fold. The ensemble approach B that spawns one convolutional neural network for each output variable reaches significantly better precision regarding stride length and width while the corresponding mean accuracies also exceed those achieved by model A. On the remaining three parameters foot angle, {heel} and toe contact time, both models perform similarly. Therefore, we consider the ensemble approach B to be superior in this context.

\begin{figure}[!h]
		\centering
		\includegraphics[width=.92\columnwidth]{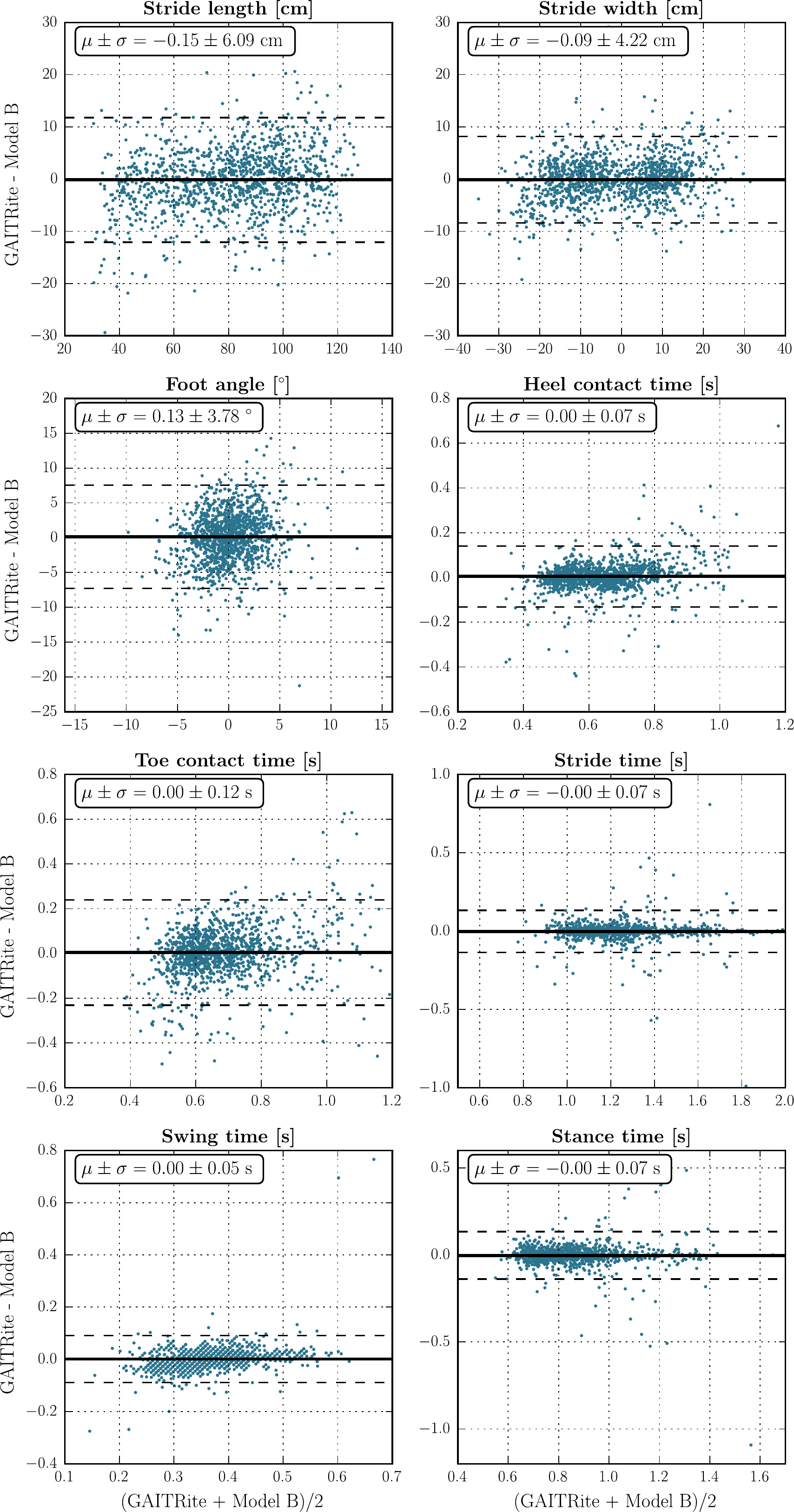}
		\caption{Bland-Altman plots for each of the output variables estimated by model B {on unseen test data}. Additionally, the mean accuracy (solid line) and ${\pm1.96\sigma}$ bounds (dashed lines) are shown.}
		\label{fig:ba-modelB}
\end{figure}

Detailed results for all output parameters on the dataset and the superior model B can be found in Fig. \ref{fig:ba-modelB} and {TABLE} \ref{tab:results-modelB}. Fig. \ref{fig:ba-modelB} includes Bland-Altman plots for each of the output variables as estimated by model B as well as the achieved mean accuracy and precision. {TABLE} \ref{tab:results-modelB} lists the error statistics for each output parameter as well as state-of-the-art results. 

\begin{table*}[!t]
	\caption{{Average estimates as achieved by model B on unseen test data and the reference system GAITRite. The error of measurement is reported as mean accuracy and precision. Additionally, results from state-of-the-art double-integration approaches are listed.}}
	\centering
	\scalebox{\tablescale}{
	\begin{tabular}{llrrrl}
		\toprule
		 &  &   \multicolumn{2}{c}{Average estimates}&  \multicolumn{2}{c}{Mean acc. $\pm$ prec.}  
		 \\\cmidrule(l){3-4}\cmidrule(l){5-6}
		Output Parameter&Unit&\multicolumn{1}{c}{Model B}&\multicolumn{1}{c}{GAITRite}&\multicolumn{1}{c}{Model B}&\multicolumn{1}{c}{State-of-the-art}\\\toprule
		
		Stride length & cm& 80.78 $\pm$ \hspace{0em} 21.82 & 80.63 $\pm$ \hspace{0em} 23.23  &-0.15 $\pm$  6.09 &-0.26 $\pm$ 8.37 $^{\text{\cite{Rampp2015}}}$\\
		&&&&&-0.16 $\pm$ 7.02 $^{\text{\cite{Ferrari2015}},\ast}$\\
		&&&&&\hspace{.33em}0.10 $\pm$ 1.90 $^{\text{\cite{Trojaniello2014}},\ast}$\\
		
		Stride width & cm& {$-1.34 \pm \hspace{0em}  12.49$} & {$-1.44 \pm \hspace{0em}  13.29$}  &  -0.09 $\pm$  4.22&\null\hspace{2.45em}--$^\text{\tiny(see text)}$  \\ 
		Foot angle & $^\circ$& -0.06 $\pm$ \hspace{.5em}  2.90 &  0.07 $\pm$ \hspace{.5em}  3.49  &  0.13 $\pm$  3.78 &\hspace{.33em}0.12 $\pm$ 3.59 $^{\text{\cite{Mariani2013}},\ast}$\\\midrule
		Heel contact time & s&  0.64 $\pm$ \hspace{.5em}  0.12 &  0.64 $\pm$ \hspace{.5em}  0.14  &  0.00 $\pm$  0.07 &\null\hspace{2.45em}--\\ 
		Toe contact time & s&  0.68 $\pm$ \hspace{.5em}  0.14 &  0.69 $\pm$ \hspace{.5em}  0.17  &  0.00 $\pm$  0.12 &\null\hspace{2.45em}--\\ 
		Stride time & s&  1.23 $\pm$ \hspace{.5em}  0.19 &  1.23 $\pm$ \hspace{.5em}  0.19  & -0.00 $\pm$  0.07 &\hspace{.33em}0.00 $\pm$ 0.07$^{\text{\cite{Rampp2015}}}$\\ 
		Swing time & s&  0.37 $\pm$ \hspace{.5em}  0.07 &  0.37 $\pm$ \hspace{.5em}  0.08  &  0.00 $\pm$  0.05 &-0.00 $\pm$ 0.05$^{\text{\cite{Rampp2015}}}$ \\ 
		Stance time & s&  0.86 $\pm$ \hspace{.5em}  0.16 &  0.85 $\pm$ \hspace{.5em}  0.16  & -0.00 $\pm$  0.07 &\hspace{.33em}0.00 $\pm$ 0.07$^{\text{\cite{Rampp2015}}}$ \\ 
		\bottomrule
		\multicolumn{6}{l}{\hfill\scriptsize Model B: $n=99$ geriatric patients and 1185 individual strides; $^\ast $  different evaluation dataset}\\
	\end{tabular}
	}
	\label{tab:results-modelB}

\end{table*}

\section{Discussion}
We present a method for data to knowledge translation in the context of sensor-based gait analysis to extract a total of 8 spatio-temporal stride characteristics. Within the proposed framework, we compare two different approaches in the case of vectorial knowledge: A) A combined modelling approach estimating the complete set of output parameters and B) An ensemble approach where individual, less complex models are spawned for each output parameter. The resulting model complexities are thereby designed to be of similar magnitude. 
With the superior ensemble approach B, the spatial parameters stride length, stride width and foot angle are estimated with mean accuracy and precision of $-0.15 \pm 6.09$ cm, $-0.09 \pm 4.22$ cm and $0.13 \pm 3.78^\circ$ respectively. The temporal stride characteristics stride, swing and stance time are predicted with average precision of $\pm0.07$, $\pm 0.05$ and $\pm 0.07$ s respectively. Additionally, the heel and toe contact times are determined up to $\pm 0.07$ and $\pm 0.12$ s respectively. Thereby, we provide technical validation on a clinically relevant dataset {containing 1185 individual strides from 99 geriatric patients}.

The estimation of stride length is outperforming the double-integration result by Rampp et al. \cite{Rampp2015} by 2.3 cm (27\%, {statistically significant}) in precision. Compared to the deep learning approach presented by Hannink et al. \cite{Hannink2016} which estimates this parameter up to $-0.27\pm5.43$ cm based on HS$\to$HS strides, our results are slightly worse. This might be due to reduced network complexity in the model presented here (Hannink et al. \cite{Hannink2016} use twice as much kernels and biases on each of the two convolutional layers) or the fact that the underlying datasets are not completely identical (101 vs. 99 patients here due to technical reasons). Regarding results on other datasets, Ferrari et al. \cite{Ferrari2015} report a measurement error of $-0.16 \pm 7.02$ cm for the parameter stride length based on double-integration and a dataset of 1314 strides captured from ten elderly PD patients. Trojaniello et al. even reach an average accuracy and precision of $0.1\pm1.9$ cm and thereby almost resolve their reference precision of $\pm 1.3$ cm \cite{Trojaniello2014}. However, this result is only evaluated on a small dataset of 532 strides from 10 elderly PD patients and points out the limits of this comparison: The results achieved have to be seen as a function of the variability across subjects captured in the evaluation dataset. A different evaluation dataset does not necessarily ensure a fair comparison of methods. And it shows the need for a unified evaluation of stride length estimation methods presented here and in the literature (\cite{Rampp2015,Hannink2016,Ferrari2015,Trojaniello2014}) on the same, large cohort study as e.g. the publicly available dataset described by Rampp et al. \cite{Rampp2015}.

Regarding stride width, there is little related work. Horak et al. \cite{Horak2013} even consider it "{difficult to obtain with body-worn sensors}". Nevertheless, Rebula et al. \cite{Rebula2013} report an intraclass correlation coefficient (ICC) of 0.88 between a motion-capture reference and a sensor-based estimation of stride width. Our result corresponds to an ICC of 0.95 and is thus outperforming state-of-the-art double-integration methods w.r.t. stride width.

Mariani et al. \cite{Mariani2013} determine the foot or turning angle up to $0.12 \pm 3.59^\circ$ on data from 10 elderly subjects and a study protocol that included a 180$^\circ$ turn as opposed to our straight walking data. In this respect, our results are comparable. 

For the parameters stride, stance and swing time, our results are identical to Rampp et al. \cite{Rampp2015} due to identical methods. Estimation of heel and toe contact times has not been reported in literature to the best of the authors' knowledge. Sabatini et al. \cite{Sabatini2005} propose detection of heel-off and toe-on/foot-flat events by thresholding the angular velocity in the sagittal plane. However, this approach is rather heuristic and the precision regarding these events was not evaluated. Thus, our result constitutes the state-of-the-art regarding sensor-based estimation of heel and toe contact times.

Based on these contact times and HS/TO events that are detectable with state-of-the-art methods, the current work enables detection of heel-off and toe-on events {(Fig.} \ref{fig:gaitrite_parameters}{) }that don't manifest that clearly in the sensor signals. Thereby, each gait cycle can be further sub-divided into loading response, mid-stance phase, terminal stance and pre-swing as defined by Perry et al. \cite{Perry1992} and their dependency on disease state or speed could be evaluated in future work. The latter would extend the work by Hebenstreit et al. \cite{Hebenstreit2015} that was based on motion capture data to a mobile setting.

The entire processing pipeline presented in the current work is based on stride segments defined from heel-strike to heel-strike and therefore independent of the zero-velocity assumption. In everyday clinical practice and the presence of impaired gait, this assumption is easily violated and limits applicability of state-of-the-art double-integration approaches \cite{Hannink2016}. However, there are no theoretical considerations that prohibit application of the proposed method to populations experiencing severe gait disturbances as in the case of spasticity. Thereby, the proposed system is a suitable substitute for the assumption-governed double-integration techniques and could enable mobile gait analysis in these clinically critical cases.

The main limitation of the proposed method is that the resulting data to knowledge translation is as good as the knowledge base. This is because the knowledge base is used to learn the non-linear relationship between the input sensor data and the output parameters and this mapping implicitly depends on the samples collected in the knowledge base. We thus strongly stress the importance of sharing benchmark datasets within the community and create larger, community-maintained knowledge bases. 

The implementation of the framework is generalisable and flexible. Information from other wearable sensing modalities (e.g. barometric pressure) can be introduced by adding additional channels to the input signal. Application to other data to knowledge translation problems in this field can be done by exchanging the knowledge base. {TensorFlow }\cite{tensorflow2015-whitepaper} {generically supports model quantisation and lower level arithmetics that are needed for inference with deep convolutional neural networks on mobile devices. Although this is not a necessity in mobile gait analysis, where the emphasis lies on the mobility of the sensing technology, it might be needed in future work.}

Future work includes the application of the proposed framework to other data to knowledge translation problems and thereby the establishment of a generally applicable system. In this respect, especially the end of training needs to be addressed in a data-adaptive manner. {As this work did not include a rigorous exploration of the parameter space (number and dimensionality of kernels, etc.), this part is left for future work.} In the context of mobile gait analysis, individualisation or domain adaptations aspects as well as sequential modelling approaches that account for across-stride context will be investigated. Additionally, the aforementioned benchmark-evaluation of several biomechanical parameter estimation methods for a fair comparison of methods will be covered in future work. 

\FloatBarrier
\section{Acknowledgements}
This work was supported by the FAU Emerging Fields Initiative (EFIMoves). The authors would like to thank Samuel Sch\"ulein and Jens Barth for their effort in compiling the benchmark dataset as well as all participants of the study for their contributions.
\bibliographystyle{IEEEtran}
\bibliography{IEEEabrv,Hannink_et_al_SensorBasedGaitParameterExtractionWithDeepCNNs_final}

\begin{thebibliography}{10}
\providecommand{\url}[1]{#1}
\csname url@samestyle\endcsname
\providecommand{\newblock}{\relax}
\providecommand{\bibinfo}[2]{#2}
\providecommand{\BIBentrySTDinterwordspacing}{\spaceskip=0pt\relax}
\providecommand{\BIBentryALTinterwordstretchfactor}{4}
\providecommand{\BIBentryALTinterwordspacing}{\spaceskip=\fontdimen2\font plus
\BIBentryALTinterwordstretchfactor\fontdimen3\font minus
  \fontdimen4\font\relax}
\providecommand{\BIBforeignlanguage}[2]{{%
\expandafter\ifx\csname l@#1\endcsname\relax
\typeout{** WARNING: IEEEtran.bst: No hyphenation pattern has been}%
\typeout{** loaded for the language `#1'. Using the pattern for}%
\typeout{** the default language instead.}%
\else
\language=\csname l@#1\endcsname
\fi
#2}}
\providecommand{\BIBdecl}{\relax}
\BIBdecl

\bibitem{Ellis2011}
T.~Ellis, J.~T. Cavanaugh, G.~M. Earhart, M.~P. Ford, K.~B. Foreman
  \emph{et~al.}, ``{Which measures of physical function and motor impairment
  best predict quality of life in Parkinson's disease?}'' \emph{Parkinsonism
  and Related Disorders}, vol.~17, no.~9, pp. 693--697, 2011.

\bibitem{Givon2009}
U.~Givon, G.~Zeilig, and A.~Achiron, ``{Gait analysis in multiple sclerosis:
  characterization of temporal-spatial parameters using GAITRite functional
  ambulation system.}'' \emph{Gait \& Posture}, vol.~29, no.~1, pp. 138--42,
  2009.

\bibitem{Webster2005}
K.~E. Webster, J.~E. Wittwer, and J.~a. Feller, ``{Validity of the
  GAITRite\textsuperscript{\textregistered} walkway system for the measurement
  of averaged and individual step parameters of gait},'' \emph{Gait \&
  Posture}, vol.~22, no.~4, pp. 317--321, 2005.

\bibitem{Kressig2004}
R.~W. Kressig, R.~J. Gregor, A.~Oliver, D.~Waddell, W.~Smith \emph{et~al.},
  ``{Temporal and spatial features of gait in older adults transitioning to
  frailty.}'' \emph{Gait \& Posture}, vol.~20, no.~1, pp. 30--5, Aug. 2004.

\bibitem{Klucken2013}
J.~Klucken, J.~Barth, P.~Kugler, J.~Schlachetzki, T.~Henze \emph{et~al.},
  ``{Unbiased and Mobile Gait Analysis Detects Motor Impairment in Parkinson's
  Disease},'' \emph{PLoS ONE}, vol.~8, no.~2, 2013.

\bibitem{Mariani2013}
B.~Mariani, M.~C. Jim\'{e}nez, F.~J.~G. Vingerhoets, and K.~Aminian, ``{On-shoe
  wearable sensors for gait and turning assessment of patients with parkinson's
  disease},'' \emph{IEEE Transactions on Biomedical Engineering}, vol.~60,
  no.~1, pp. 155--158, 2013.

\bibitem{Rampp2015}
A.~Rampp, J.~Barth, S.~Sch\"{u}lein, K.-G. Ga\ss{}mann, J.~Klucken
  \emph{et~al.}, ``{Inertial Sensor Based Stride Parameter Calculation from
  Gait Sequences in Geriatric Patients.}'' \emph{IEEE Transactions on
  Biomedical Engineering}, vol.~62, no.~4, pp. 1089--1097, 2014.

\bibitem{Ferrari2015}
A.~Ferrari, P.~Ginis, M.~Hardegger, F.~Casamassima, L.~Rocchi \emph{et~al.},
  ``{A Mobile Kalman-Filter Based Solution for the Real-Time Estimation of
  Spatio-Temporal Gait Parameters},'' \emph{IEEE Transactions on Neural Systems
  and Rehabilitation Engineering}, no.~99, 2015.

\bibitem{Trojaniello2014}
D.~Trojaniello, A.~Cereatti, E.~Pelosin, L.~Avanzino, A.~Mirelman
  \emph{et~al.}, ``{Estimation of step-by-step spatio-temporal parameters of
  normal and impaired gait using shank-mounted magneto-inertial sensors:
  application to elderly, hemiparetic, parkinsonian and choreic gait},''
  \emph{Journal of Neuroengineering and Rehabilitation}, vol.~11, no.~1, p.
  152, 2014.

\bibitem{Aminian2002}
K.~Aminian, B.~Najafi, C.~B\"{u}la, P.~F. Leyvraz, and P.~Robert,
  ``{Spatio-temporal parameters of gait measured by an ambulatory system using
  miniature gyroscopes},'' \emph{Journal of Biomechanics}, vol.~35, no.~5, pp.
  689--699, 2002.

\bibitem{Salarian2013}
A.~Salarian, P.~R. Burkhard, B.~M. Jolles, and K.~Aminian, ``{A Novel Approach
  to Reducing Number of Sensing Units for Wearable Gait Analysis Systems},''
  \emph{IEEE Transactions on Biomedical Engineering}, vol.~60, no.~1, pp.
  72--77, 2013.

\bibitem{Rebula2013}
J.~R. Rebula, L.~V. Ojeda, P.~G. Adamczyk, and A.~D. Kuo, ``{Measurement of
  foot placement and its variability with inertial sensors.}'' \emph{Gait \&
  Posture}, vol.~38, no.~4, pp. 974--80, Sep. 2013.

\bibitem{Hannink2016}
J.~Hannink, T.~Kautz, C.~F. Pasluosta, J.~Barth, S.~Sch\"ulein \emph{et~al.},
  ``{Stride Length Estimation with Deep Learning},'' \emph{{IEEE Transactions
  on Neural Systems and Rehabilitation}}, 2016, submitted.

\bibitem{He2015}
K.~He, X.~Zhang, S.~Ren, and J.~Sun, ``{Deep Residual Learning for Image
  Recognition},'' \emph{arXiv preprint}, 2015,
  \url{https://arxiv.org/abs/1512.03385}.

\bibitem{Vinyals2015}
O.~Vinyals, A.~Toshev, S.~Bengio, and D.~Erhan, ``{Show and Tell: A Neural
  Image Caption Generator},'' in \emph{The IEEE Conference on Computer Vision
  and Pattern Recognition (CVPR)}, 2015.

\bibitem{Taigman2014}
Y.~Taigman;, M.~Yang;, M.~Ranzato, and L.~Wolf, ``{DeepFace: Closing the Gap to
  Human-Level Performance in Face Verification},'' in \emph{The IEEE Conference
  on Computer Vision and Pattern Recognition (CVPR)}, 2014.

\bibitem{Hammerla2015}
N.~Y. Hammerla, J.~M. Fisher, P.~Andras, L.~Rochester, R.~Walker \emph{et~al.},
  ``{PD Disease State Assessment in Naturalistic Environments using Deep
  Learning},'' in \emph{Conference on Innovative Applications of Artificial
  Intelligence}, 2015.

\bibitem{Hammerla2016}
N.~Y. Hammerla, S.~Halloran, and T.~Ploetz, ``{Deep, Convolutional, and
  Recurrent Models for Human Activity Recognition using Wearables},''
  \emph{arXiv}, 2016, \url{http://arxiv.org/abs/1604.08880}.

\bibitem{Ordonez2016}
F.~J. Ord\'{o}\~{n}ez and D.~Roggen, ``{Deep convolutional and LSTM recurrent
  neural networks for multimodal wearable activity recognition},''
  \emph{Sensors}, vol.~16, no.~1, 2016.

\bibitem{Zheng2014}
Y.~Zheng, Q.~Liu, E.~Chen, Y.~Ge, and J.~L. Zhao, ``{Time series classification
  using multi-channels deep convolutional neural networks},'' in \emph{Lecture
  Notes in Computer Science}, vol. 8485, 2014, pp. 298--310.

\bibitem{Lane2015}
N.~D. Lane and P.~Georgiev, ``{Can Deep Learning Revolutionize Mobile
  Sensing?}'' \emph{Proceedings of the 16th International Workshop on Mobile
  Computing Systems and Applications - HotMobile '15}, pp. 117--122, 2015.

\bibitem{Burns2010}
A.~Burns, B.~R. Greene, M.~J. McGrath, T.~J. O'Shea, B.~Kuris \emph{et~al.},
  ``{SHIMMER\textsuperscript{TM} -- A wireless sensor platform for noninvasive
  biomedical research},'' \emph{Sensors}, vol.~10, no.~9, pp. 1527--1534, 2010.

\bibitem{Menant2009}
J.~C. Menant, J.~R. Steele, H.~B. Menz, B.~J. Munro, and S.~R. Lord, ``{Effects
  of walking surfaces and footwear on temporo-spatial gait parameters in young
  and older people.}'' \emph{Gait \& Posture}, vol.~29, no.~3, pp. 392--7, Apr.
  2009.

\bibitem{Salzman2010}
B.~Salzman, ``{Gait and Balance Disorders in Older Adults},'' \emph{American
  Family Physician}, vol.~82, no.~1, pp. 61--68, 2010.

\bibitem{Glorot2011}
X.~Glorot, A.~Bordes, and Y.~Bengio, ``{Deep Sparse Rectifier Neural
  Networks},'' \emph{AISTATS}, vol.~15, pp. 315--323, 2011.

\bibitem{Boureau2010}
Y.~Boureau, J.~Ponce, and Y.~LeCun, ``{A theoretical analysis of feature
  pooling in visual recognition},'' in \emph{{Proc. of the 27th International
  Conference on Machine Learning (ICML-10)}}, 2010, pp. 111--118.

\bibitem{Goodfellow2016}
I.~Goodfellow, Y.~Bengio, and A.~Courville, ``Deep learning,'' 2016, book in
  preparation for MIT Press, available at
  \url{http://www.deeplearningbook.org}.

\bibitem{Kingma2015}
D.~P. Kingma and J.~L. Ba, ``{Adam: a Method for Stochastic Optimization},''
  \emph{International Conference on Learning Representations}, pp. 1--13, 2015.

\bibitem{Srivastava2014}
N.~Srivastava, G.~Hinton, A.~Krizhevsky, I.~Sutskever, and R.~Salakhutdinov,
  ``{Dropout : A Simple Way to Prevent Neural Networks from Overfitting},''
  \emph{Journal of Machine Learning Research}, vol.~15, pp. 1929--1958, 2014.

\bibitem{tensorflow2015-whitepaper}
M.~Abadi, A.~Agarwal, P.~Barham, E.~Brevdo, Z.~Chen \emph{et~al.},
  ``{TensorFlow}: Large-scale machine learning on heterogeneous systems,''
  2015, software available at \url{https://tensorflow.org}.

\bibitem{Horak2013}
F.~B. Horak and M.~Mancini, ``{Objective biomarkers of balance and gait for
  Parkinson's disease using body-worn sensors},'' \emph{Movement Disorders},
  vol.~28, no.~11, pp. 1544--1551, 2013.

\bibitem{Sabatini2005}
A.~M. Sabatini, ``{Quaternion-based strap-down integration method for
  applications of inertial sensing to gait analysis},'' \emph{Medical and
  Biological Engineering and Computing}, vol.~43, no.~1, pp. 94--101, 2005.

\bibitem{Perry1992}
J.~Perry, J.~M. Burnfield, and L.~M. Cabico, \emph{{Gait Analysis: Normal and
  Pathological Function}}.\hskip 1em plus 0.5em minus 0.4em\relax Slack
  Thorofare, NJ, 1992, vol.~12.

\bibitem{Hebenstreit2015}
F.~Hebenstreit, A.~Leibold, S.~Krinner, G.~Welsch, M.~Lochmann \emph{et~al.},
  ``{Effect of walking speed on gait sub phase durations},'' \emph{Human
  Movement Science}, vol.~43, pp. 118--124, 2015.

\end{thebibliography}

\end{document}